\documentclass[10pt,twocolumn,letterpaper]{article}

\usepackage{cvpr}              

\usepackage[dvipsnames]{xcolor}

\definecolor{cvprblue}{rgb}{0.21,0.49,0.74}
\usepackage[pagebackref,breaklinks,colorlinks,citecolor=cvprblue]{hyperref}

\usepackage[bb=boondox]{mathalfa}
\usepackage{relsize}
\usepackage{graphicx}
\usepackage{listings}

\def\confName{CVPR}
\def\confYear{2024}

\title{Attention Deficit is Ordered! Fooling Deformable Vision Transformers with Collaborative Adversarial Patches}

\author{Quazi Mishkatul Alam\\
University of California, Riverside\\
Riverside, CA, USA\\
{\tt\small qalam001@ucr.edu}
\and
Bilel Tarchoun\\
Ecole Nationale d'Ingénieurs de Sousse\\
Sousse, Tunisia\\
{\tt\small bilel.tarchoun@eniso.u-sousse.tn}
\and
Ihsen Alouani\\
Queen's University Belfast\\
Belfast, UK\\
{\tt\small i.alouani@qub.ac.uk}
\and
Nael Abu-Ghazaleh\\
University of California, Riverside\\
Riverside, CA, USA\\
{\tt\small naelag@ucr.edu}}

\begin{document}
\maketitle
\begin{abstract}
The latest generation of transformer-based vision models has proven to be superior to Convolutional Neural Network (CNN)-based models across several vision tasks, largely attributed to their remarkable prowess in relation modeling. Deformable vision transformers significantly reduce the quadratic complexity of attention modeling by using sparse attention structures, enabling them to incorporate features across different scales and be used in large-scale applications, such as multi-view vision systems. Recent work has demonstrated adversarial attacks against conventional vision transformers; we show that these attacks do not transfer to deformable transformers due to their sparse attention structure. Specifically, attention in deformable transformers is modeled using pointers to the most relevant other tokens. In this work, we contribute for the first time adversarial attacks that manipulate the attention of deformable transformers, redirecting it to focus on irrelevant parts of the image. We also develop new collaborative attacks where a source patch manipulates attention to point to a target patch, which contains the adversarial noise to fool the model. In our experiments, we observe that altering less than 1\% of the patched area in the input field results in a complete drop to 0\% AP in single-view object detection using MS COCO and a 0\% MODA in multi-view object detection using Wildtrack.
\end{abstract}    
\section{Introduction}
\label{sec:intro}

Since the early development of machine learning-based vision models, their susceptibility to adversarial attacks has been a persistent security concern \cite{akhtar2018threat}. Currently, the best-performing models use transformer architectures due to their superior performance relative to CNN-based architectures. Vision transformers leverage a global attention mechanism to learn and exploit the spatial relationships within an image or video frame. In addition to their superior performance, they were also conjectured to have robustness advantages with respect to adversarial attacks~\cite{aldahdooh2021reveal, benz2021adversarial}, with studies demonstrating that attacks on traditional convolutional vision models do not transfer to transformers~\cite{mahmood2021robustness,bai2021transformers}. The basis of this conjecture lies in the drastically different architecture of transformers, which use stacked layers of self-attention modules. These modules excel in learning and exploiting long-range information, making them resilient to the high-frequency nature of CNN-based adversarial attacks~\cite{shao2021adversarial}.

\begin{figure*}[ht]
  \centering
  \includegraphics[width=0.78\linewidth]{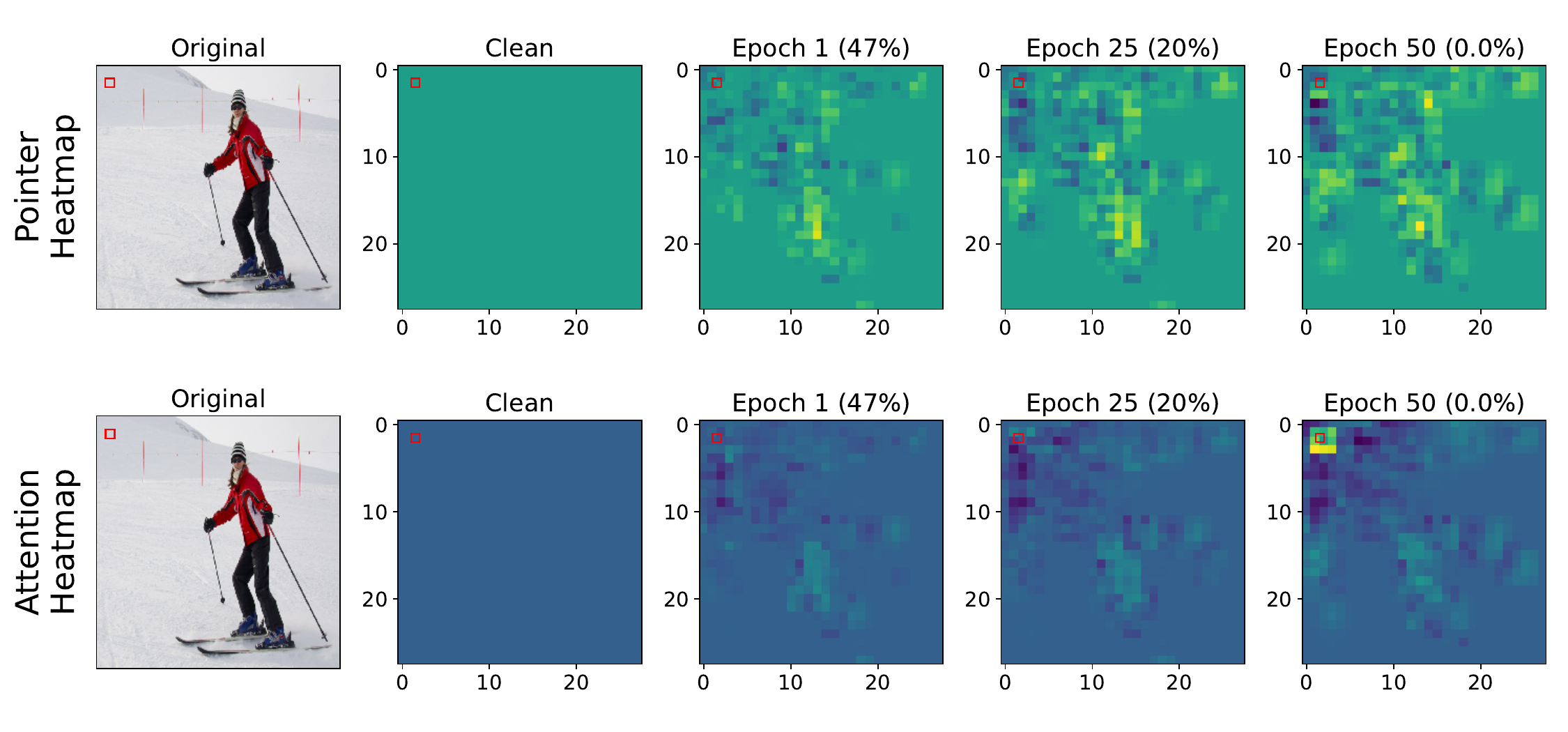}
  \caption{Attention Deficit Attack: The top row of the figure displays heatmaps of the attention pointers directed at the target area (marked in a red box). As the attack progresses, an increasing number of pointers are adversarially redirected to the target. The bottom row displays heatmaps of the attention scores (magnitude), where there is also a gradual increase in the attention scores directed towards the target.}
  \label{fig:heatmap}
\end{figure*}


Subsequent works have demonstrated that new attacks can also be developed to compromise Vision Transformers (ViTs), particularly targeting their dot-product-based attention module~\cite{lovisotto2022give,fu2022patch}. Within this module, each layer comprises multiple attention heads, each of which is dedicated to modeling the spatial relationships within an image. The aforementioned attacks craft adversarial patches that increase the attention of all query tokens towards the malicious key tokens forming the patch, thereby exaggerating its impact to the attacker's advantage. These works have illustrated that carefully designed loss functions targeting the dot-product attention in transformers, can lead to adversarial effects.


Transformers face limitations in run-time scalability due to the complexity of the multi-headed dot-product attention module. In its design, attentions are learned at the granularity of individual pixels, with each head potentially operating at multiple resolutions, resulting in a large number of tokens. This dense attention module make transformers difficult to scale; for instance, the dot-product self-attention module exhibits quadratic time complexity in relation to the number of tokens (e.g., pixels). To enhance the computational performance and scalability of transformers, deformable transformers utilize a sparsified attention module~\cite{zhu2020deformable}. Specifically, rather than a query token attending to all possible key tokens, deformable transformers determine a fixed-size set of key tokens (for example, 4) that are most closely associated with the query token, based on its features. Learning this attention involves identifying these tokens and learning the attention score for each of them. Moreover, Deformable Vision Transformers (DViTs) are inherently designed to consider features at multiple scales, enabling them to capture objects of varying sizes, from very small to large. DViTs combine the benefits of global attention with reasonable run-time properties by focusing their attentions on a small subset of tokens, achieving significant speedup over ViTs with excellent model performance~\cite{zhu2020deformable}. DViTs exhibit linear scaling (as opposed to quadratic) with respect to the number of tokens \cite{zhu2020deformable}. They are also notably faster to train, requiring on average 10x fewer epochs to match the performance of ViTs ~\cite{zhu2020deformable}. These advantages have made DViTs a popular choice in vision applications.

\begin{figure}[t]
    \centering
    \includegraphics[width=0.9\columnwidth]{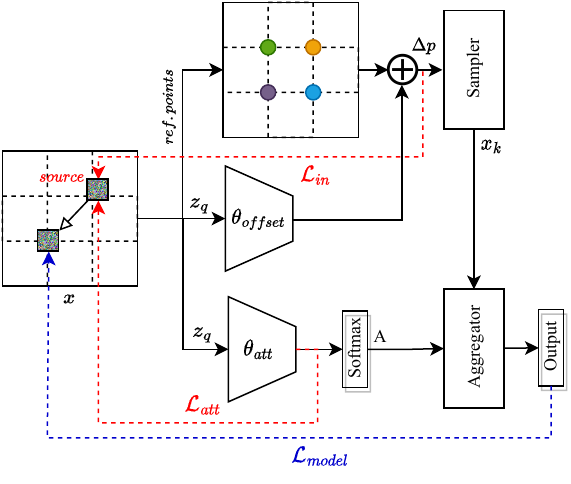}
    \caption{An overview of the proposed Collaborative Patch attack on DViTs, which is one of our four pointer-based attack strategies. All attacks involve the manipulation of pointers through an adversarial patch. This patch is generated by backpropagating the pointer loss, which identifies the most relevant tokens, and the attention loss (both illustrated in red). These components collectively enable the attacker to direct attention to a strategically crafted target patch, designed to maximize the model loss (blue).}
    \label{fig:illustration}
\end{figure}

In this paper, we explore for the first time the adversarial properties of DViTs. We first show that the attacks on conventional vision transformers do not successfully transfer to DViTs, we suspect because of their different attention model. We contribute new adversarial attacks catered towards manipulating the attention directions as well as the scores in the sparse attention module used by DViTs. A illustration of the state of the sparse attention module as the construction of the patch is in progress, is shown in Fig.~\ref{fig:heatmap}; as the attack progresses left to right in the figure, more of the attention pointers are redirected to the target with higher attention scores. Since the sparse attention has two components (i.e., a direction and a score), in subsequent sections, we will refer to them as \emph{pointers} and \emph{attentions}, respectively. Fig.~\ref{fig:illustration} overviews how the attack works.

Our attacks uniquely decouple the attention manipulation by backpropagating the pointer and attention loss functions directly. This approach enables the manipulation of pointers to target an attacker-controlled set of locations on the image. Leveraging this insight, we propose a family of four distinct pointer-based attacks against DViTs: (i) Inward Pointer, (ii) Outward Pointer, (iii) Standalone Patch, and (iv) Collaborative Patch, each with its own specific attack objectives. Most notably, in our collaborative patch attack, we demonstrate that two colluding patches — one patch redirecting the pointers towards the other patch, which contains the adversarial noise to fool the model — can synergize to create a highly effective attack.

In summary, we make the following contributions:
\begin{itemize}
    \item For the first time, we develop attacks against Deformable Vision Transformers (DViTs), which are resilient against existing attention-based attacks. Our attacks are designed to exploit the unique sparse attention mechanism inherent in such models.
    \item We introduce four distinct attack strategies, each with its unique objective. This includes a novel collaborative attack that leverages patch cooperation.
    \item We evaluate the effectiveness of these attacks on the MS COCO \cite{lin2014microsoft} dataset and the multi-view object detection dataset Wildtrack \cite{chavdarova2018wildtrack}. Our attacks exhibit excellent effectiveness against both Deformable DeTr (single-view) and MVDeTr (multi-view) models.
\end{itemize}

\section{Background and Preliminaries}
\label{sec:preliminaries}

\subsection{Multi-headed Dense Dot-product Attention}
Transformer architectures~\cite{vaswani2017attention} were initially designed for modeling complex contextual relations among word tokens in textual data. Subsequently, these architectures have been adapted for vision, where they achieved notable improvements in performance across a range of vision tasks~\cite{khan-22}. In Vision Transformers (ViT), the initial stages of the processing pipeline still utilize convolutional layers for feature extraction, as they excel at perceiving local features. In practice, a VGG \cite{simonyan2014very} or ResNet \cite{he2016deep} based feature extractor is used to convert the input frames into feature maps. A transformer-based module, consisting of an encoder and a decoder, each utilizing multi-headed attention mechanisms, subsequently processes these feature maps~\cite{vaswani2017attention,khan-22}. The transformer enables the model to learn long-range spatial relations within the image or video frame, resulting in superior performance compared to traditional vision models. 

The multi-headed dot-product attention module is responsible for capturing the relations among the tokens in the input data, represented as a collection of queries and keys. Both the encoder and the decoder consist of stacked layers of these attention modules that calculate self-attention and cross-attention among input queries and keys respectively, at multiple heads. Given a collection of queries and keys, each having sizes $Q$ and $K$ respectively, if $z_q \in \mathbb{R}^C$ represents the feature vector of the $q^{th}$ query and $x_k \in \mathbb{R}^C$ represents the feature vector of the $k^{th}$ key, then the multi-headed attention function can be formulated as follows~\cite{vaswani2017attention}:
\begin{equation}
    Att(z_q, x) = \sum_{h=1}^{H} W_h \Bigg[\sum_{k=1}^{K} A_{hqk} . W'_h x_k\Bigg]
\end{equation}
Where H is the total number of heads, and $W_h \in \mathbb{R}^{C \times C_h}$ and $W'_h \in \mathbb{R}^{C_h \times C}$ are learnable weights in each head, and $C$ and $C_h$ are the total and per-head lengths of the feature vectors respectively. The attention weights are calculated as: $A_{hqk} \propto \exp{\frac{z_q^T U_h^T U'_h x_k}{\sqrt{C_h}}}$, which is normalized over all key elements $\sum_{k \in K} A_{hqk} = 1$, where $U_h, U'_h \in \mathbb{R}^{C_h \times C}$ are also learnable weights. This multi-headed dot-product attention module calculates self-attention across the feature map at the encoder. Similarly, at the decoder, it evaluates self-attention to establish connections among object queries, and cross-attention to identify the relationships between object queries and the encoded features.

\subsection{Multi-headed Sparse Deformable Attention} 
A major drawback of the multi-headed dot-product attention module is the computational cost associated with calculating the attention weights $A$. In large feature spaces, such as features considered at multiple scales for identifying objects of varying sizes, or features from multiple cameras, computing the attention weights can become quite expensive. Moreover, convergence is also an issue, as the attention weights for all keys are initialized similarly (uniformly or randomly) regardless of their relevance to the queries. 

To address the computational challenges in ViTs, Deformable Vision Transformers (DViTs) were proposed. DViTs replace the dot-product attention module, which tracks the relationship between all pairs of tokens in the attention matrix, with sparsified deformable attention mechanisms. In this mechanism, rather than learning attention weights for every key in relation to a query, attention is focused on a considerably smaller subset of keys. This approach of sparse attention incorporates a new learnable component that, for each query, predicts the pointers (offsets from a reference point on a 2D plane) and the corresponding attention scores for a select few prominent keys. In future sections, this component is referred to as the \emph{pointer predictor} module. In such deformable architectures, attention is no longer a single scalar value; instead, it takes the form of a vector that includes both a direction and a magnitude. The multi-headed deformable attention \cite{zhu2020deformable} function can be formulated as:
\begin{equation}
    DfAtt(z_q, p_q, x) = \sum_{h=1}^{H} W_h \Bigg[\sum_{k=1}^{R} A_{hqk} . W'_h  x(p_q + \Delta p_{hqk})\Bigg]
\end{equation}
Where $R \ll K$ represents the maximum number of salient keys, $p_q$ symbolizes the reference point for the $q^{th}$ query, and $\Delta p_{hqk}$ denotes - for the $q^{th}$ query - the $k^{th}$ offset at the $h^{th}$ attention head. The pointers and the attention weights in DViTs are computed as: $\Delta p_{hqk} = U_h z_{hq}$ and $A_{hqk} = U'_h z_{hq}$, respectively, where $z_{hq} \in \mathbb{R}^{C_h}$ represents the per-head feature vectors. Furthermore, $U_h \in \mathbb{R}^{2R \times C_h}$ and $U'_h \in \mathbb{R}^{R \times C_h}$ denote the additional learnable weights in DViTs.

\subsection{Adversarial Attacks on Transformers} 
Despite initial beliefs that transformers have built-in resistance to adversarial attacks, researchers have since devised new attacks specifically aimed at the multi-headed dot-product attention module within these models. These emerging attacks \cite{lovisotto2022give, fu2022patch} manipulate the attention weights $A$ of the dot-product attention module. They employ a localized adversarial patch with unbounded noise to shift a key feature towards the centroid of the query features by maximizing the following per-head loss function:
\begin{equation}
    \mathcal{L}_{h} = \frac{1}{Q}\sum_{q=1}^{Q} A_{hqi^*}
\end{equation}
where $i^*$ represents the target key. This per-head loss can be further extended to all the layers in the encoder and decoder to form a total loss function: $\mathcal{L} = \sum_{l=1}^{L} \mathcal{L}_h$.

We demonstrate that these types of attacks are ineffective against deformable transformers for the following reasons: (i) the query tokens in deformable transformers do not attend to all key tokens; therefore, even for the most popular keys, there will be queries that are not associated with them, and (ii) the keys that each query attends to are dynamic and depend on the query's feature vector, lacking a consistent pattern that can be exploited for these attacks.


\section{Adversarial Attacks on DViTs}
\label{sec:method}

We propose, to our knowledge, the first adversarial attacks on Deformable Vision Transformers (DViTs). Our attacks utilize adversarial perturbations to manipulate the pointers and their associated attentions, which are at the core of attention modeling within DViTs. Once attackers gain control over these elements, they can exploit this capability to implement a number of flexible attack strategies. We explore a family of four pointer-based attacks, each characterized by the direction of the pointers and the objective of the patches. We anticipate that future research may explore additional strategies. The four attacks we present are: (i) Inward Pointer (IP) attack, where pointers are manipulated to converge towards predetermined target point(s); (ii) Outward Pointer (OP) attack, where pointers are manipulated to diverge from predetermined target point(s); (iii) Standalone Patch (SP) attack, where each patch draws the pointers to itself, while also attacking the model loss; and (iv) Collaborative Patch (CP) attack, where source patches manipulate pointers to target patches that attack the model loss.

In each of these attack scenarios, we assume white-box access to the transformer’s parameters. Specifically, we require access to the \emph{pointer predictor} module that determines the pointers and their corresponding attentions. While it is possible that black-box attacks could be developed using proxy models, we do not explore this possibility in this paper. We confine the patches to a small area within the image without limiting the noise budget within the patch, consistent with other physical patch-based attacks \cite{lovisotto2022give, fu2022patch}. Attacks (i) through (iii) can be performed in real-life by printing a patched T-shirt and walking in front of a camera. Whereas, the attack (iv) requires a static patch, such as, a patched bag or a tent in addition to the printed T-shirt. Lastly, for the attacks (iii) and (iv) we further assume knowledge of the loss function $\mathcal{L}_{model}$ used in the training of the end-to-end model so that it can be utilized during the adversarial patch generation.

\subsection{Manipulating Pointer Direction}
We begin by describing the IP and OP attacks, where we solely manipulate the pointers of each token to make them either converge towards or diverge away from a set of points. These attacks demonstrate the feasibility of attacking DViTs without access to the model loss $\mathcal{L}_{model}$ used in their training. For these attacks, we have the following scenario: we have a collection of input samples $X = [x_1, \ldots, x_N]$, where $x \in \mathbb{R}^{3 \times H \times W}$, and possess white-box access to the weights of the DViT. However, during the adversarial patch generation, we do not have the ground-truth labels $Y = [y_1, \ldots, y_N]$, nor any information about the model loss $\mathcal{L}_{model}$. Furthermore, we choose a collection of source locations $S = [s_1, s_2, \ldots]$ for positioning the source patches, and a collection of target locations $T = [t_1, t_2, \ldots]$, to and from which pointers either converge or diverge respectively. The number of these source and target patches should not exceed the number of pointers $R$ in each head of the sparse attention module. \\

\noindent\textbf{Inward Pointer (IP) Attack:} In this attack, our goal is to redirect the pointers of all tokens to converge towards the target locations in $T$, and also to maximize their associated attentions. We extend $T$ by repeating locations as necessary, after ensuring that each location is included at least once, thereby making its size equal to $R$. We also utilize a perturbation frame $E \in \mathbb{R}^{3 \times H \times W}$ and a mask $M \in (\mathbb{0}, \mathbb{1})^{H \times W}$ to channel gradients to the appropriate locations during optimization through backpropagation. Given a location $(p, q)$ and a length $a$, we use the following function to set an entry $(i, j)$ in mask $M$:  
\begin{equation}
    M_{ij}^{pqa} = \begin{cases}
      1, & \text{if } p \leq i \leq p + a \text{ and } q \leq j < q + a \\
      0, & \text{otherwise}
    \end{cases}
    \label{eqn:mask_identity}
\end{equation}
Using this function, we initialize mask $M$ for locations $(p, q) \in S$, and patch-length $a$ as follows:
\begin{equation}
    \forall_{(p, q) \in S}\forall_{i \in W , j \in H} M_{ij}^{pqa} 
    \label{eqn:mask_set}
\end{equation}
We use the perturbation frame $E$ and mask $M$ to perturb the sample data $X$ in the following way:
\begin{equation}
    X' = X + E \otimes M
    \label{eqn:perturbed_data}
\end{equation}
Where $\otimes$ represents the element-wise matrix multiplication operator.

In the IP attack, we execute a single-task optimization that redirects the pointers towards the locations in $T$ and maximizes their corresponding attentions. To achieve this, we define the following inward pointer loss function $\mathcal{L}_{in}$ and the attention loss function $\mathcal{L}_{att}$:
\begin{equation}
    \mathcal{L}_{in} = \sum_{l=1}^{L}\sum_{h=1}^{H}\frac{1}{DQK}\sum_{q=1}^{Q}\sum_{d=1}^{D}\sum_{k=1}^{R} (p_q + \Delta p_{lhqdk} - T_k)^2 
    \label{eqn:inptr_loss}
\end{equation}
\begin{equation}
    \mathcal{L}_{att} = -\sum_{l=1}^{L}\sum_{h=1}^{H}\frac{1}{DQK}\sum_{q=1}^{Q}\sum_{d=1}^{D}\sum_{k=1}^{R} A_{lhqdk}
    \label{eqn:att_loss}
\end{equation}
where $A$ represents the pre-softmax attentions in the sparse attention module. Then, we use the combined loss function $\mathcal{L} = \mathcal{L}_{in} + \mathcal{L}_{att}$ to optimize the perturbation frame $E$ in order to generate the source patches. In this and the following attack strategies, the gradient surgery technique PCGrad \cite{yu2020gradient} is utilized to handle any conflicting gradients. \\

\noindent\textbf{Outward Pointer (OP) Attack:} 
In this attack, our goal is to redirect the pointers of all tokens to diverge from the target locations in $T$, and also to maximize their associated attentions. We achieve this by executing a single-task optimization. 
Similar to the previous attack, we extend $T$ to make its size equal to $R$. We also use a perturbation frame $E$ and a mask $M$ constructed using Eqn.~\ref{eqn:mask_identity} and Eqn.~\ref{eqn:mask_set} respectively, which is added to the sample data $X$ in accordance with Eqn.~\ref{eqn:perturbed_data}. Differing from the previous attack, we define the following outward loss function $\mathcal{L}_{out}$:
\begin{small}
\begin{equation}
    \mathcal{L}_{out} = -\sum_{l=1}^{L}\sum_{h=1}^{H}\frac{1}{DQK}\sum_{q=1}^{Q}\sum_{d=1}^{D}\sum_{k=1}^{R} (p_q + \Delta p_{lhqdk} - T_k)^2 
\end{equation}
\end{small}
Then, we use the $\mathcal{L}_{att}$ from Eqn.~\ref{eqn:att_loss}, along with the combined loss function $\mathcal{L} = \mathcal{L}_{out} + \mathcal{L}_{att}$ to optimize the perturbation frame $E$ in order to generate the source patches.


\section{Multi-objective Attacks}
In the SP and CP attacks, we not only manipulate the pointers but also adversarially affect the model loss. For these attacks, we have the following scenario: in addition to the collection of input samples $X = [x_1, \ldots, x_N]$, we also have the associated ground-truth labels $Y = [y_1, \ldots, y_N]$. Furthermore, we possess white-box access to the weights of the DViT, and have full information about the model loss $\mathcal{L}_{model}$. For the CP attack, we need both source locations $S$ and target locations $T$, while in the SP attack, the source patches merge with the target patches, necessitating only the target locations. \\

\noindent\textbf{Collaborative Patch (CP) Attack:} In this attack, our goal is twofold: (1) utilize the source patches to redirect the pointers of all tokens to converge towards the target locations in $T$, and also maximize their associated attentions; and (2) at the target locations in $T$ create adversarial target patches that affect the DViT's model loss. We achieve this by executing a multi-task optimization. To perform multiple attack objectives, alongside the source perturbation frame $E$ and source mask $M$, we introduce a target perturbation frame $F \in \mathbb{R}^{3 \times H \times W}$ and a target mask $N \in (\mathbb{0}, \mathbb{1})^{H \times W}$. The source and target masks are initialized using Eqn.~\ref{eqn:mask_identity} and Eqn.~\ref{eqn:mask_set} with the respective points in $S$ and $T$. We use the perturbation frames and the masks to perturb the sample data $X$ in the following way:
\begin{equation}
    X' = X + E \otimes M + F \otimes N
\end{equation}

\noindent We construct the source and target patches on two separate perturbation frames. We optimize the perturbation frame $E$ utilizing the  $\mathcal{L}_{in}$ from Eqn.~\ref{eqn:inptr_loss} and $\mathcal{L}_{att}$ from Eqn.~\ref{eqn:att_loss} to formulate the combined loss function $\mathcal{L} = \mathcal{L}_{in} + \mathcal{L}_{att}$ in order to generate the source patches. Furthermore, we optimize the perturbation frame $F$ utilizing the model loss as $-\mathcal{L}_{model}$ in order to generate the target patches. \\

\begin{figure*}[t]
    \centering
    \begin{minipage}{\textwidth}
        \includegraphics[width=\linewidth]{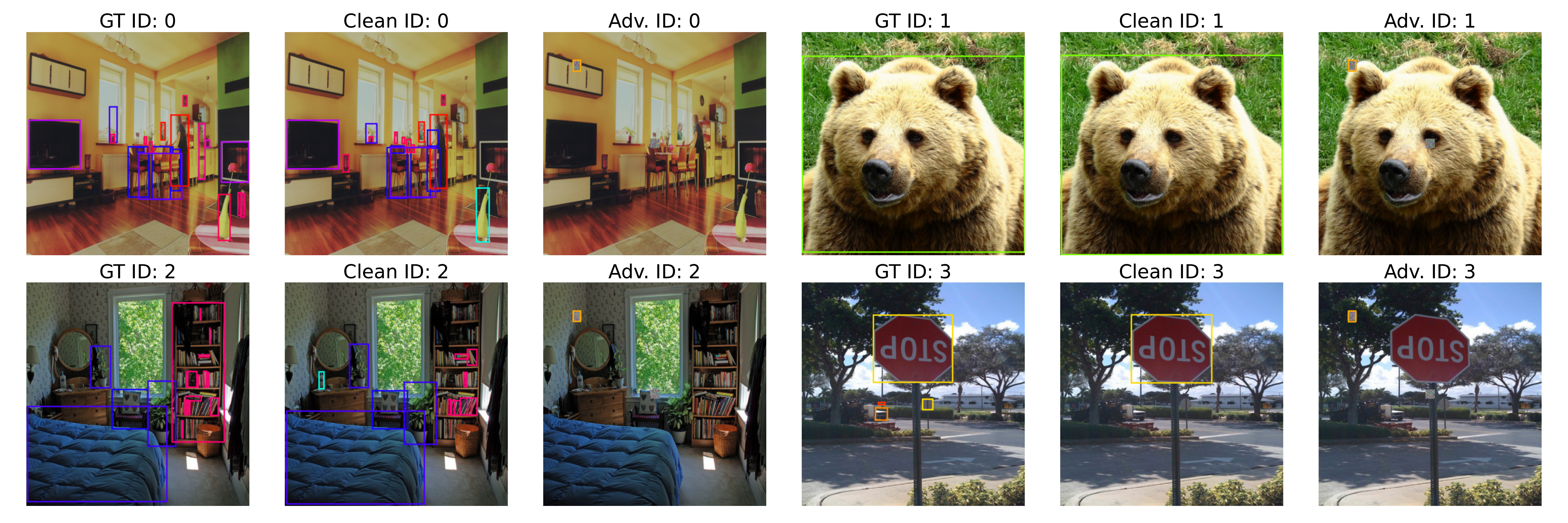}
        \caption{Bounding box comparison using the CP attack on Deformable DeTr: ground truth vs. clean samples vs. adversarial samples. Notably, the object detection performance is severely compromised (absent) in the case of adversarial samples.}
        \label{fig:bbox_CP}
    \end{minipage}
\end{figure*}

\noindent\textbf{Standalone Patch (SP) Attack:} In this attack, our goal is to create standalone patches that not only redirect the pointers of all tokens to themselves and maximize associated attentions but also maximizes the model loss. We achieve this by executing a single-task optimization. It is particularly useful in scenarios where we require only a single type of patch. Unlike the CP attack, we use only a single perturbation frame, $F \in \mathbb{R}^{3 \times H \times W}$, and a mask $N \in (\mathbb{0}, \mathbb{1})^{H \times W}$. In this attack, the source and target locations are identical; therefore, only the target locations in $T$ are utilized to initialize the mask $N$ using Eqn.~\ref{eqn:mask_identity} and Eqn.~\ref{eqn:mask_set}. We use the perturbation frame $F$ and mask $N$ to perturb the sample data $X$ in the following way:
\begin{equation}
    X' = X + F \otimes N
\end{equation}
Then, we use the $\mathcal{L}_{in}$ from Eqn.~\ref{eqn:inptr_loss} and $\mathcal{L}_{att}$ from Eqn.~\ref{eqn:att_loss}, along with the combined loss function $\mathcal{L} = \mathcal{L}_{in} + \mathcal{L}_{att} - \mathcal{L}_{model}$ in order to generate the standalone patches.

\section{Evaluation}

We evaluate the proposed adversarial attacks on two well-known deformable vision transformers (DViTs). Initially, we perform our attacks on the object detector Deformable DeTr \cite{zhu2020deformable}, utilizing the MS COCO dataset \cite{lin2014microsoft}, which was used to evaluate the original design. Besides presenting the attack's effectiveness, we further examine its dynamics in relation to various attack parameters. Subsequently, we demonstrate the applicability of our attacks in large feature spaces. In particular, we apply our attack on the Shadow transformer, a deformable transformer employed in the multi-view object detector MVDeTr \cite{hou2021multiview} for aggregating world feature maps. The attack's performance on MVDeTr is evaluated using the Wildtrack multi-view dataset. \\


\noindent\textbf{Adversarial Patch Generation.} 
For both datasets, we sampled 360 frames for training and 40 frames for evaluating robustness. Training was discontinued either when robust accuracy reached zero or after completing 100 iterations. We employed the Adam optimizer to optimize the adversarial patch using model gradients, setting the learning rate ($\eta$) to 0.22. Additionally, we utilized a scheduler with a decay rate ($\gamma$) of 0.95 and a step size of 10. Due to memory constraints, we limited the batch size for MVDeTr to one. Finally, in our evaluation, AP (Average Precision) was utilized as the performance metric for single-view object detection, and MODA (Multi-Object Detection Accuracy) for multi-view object detection, aligning with the metrics used by the authors in Detr~\cite{zhu2020deformable} and MVDeTr~\cite{hou2021multiview} respectively.


\subsection{Attack on Deformable Detr}
We evaluated our family of pointer-based attacks using the Deformable DeTr \cite{zhu2020deformable}, employing all four strategies: (i) Inward Pointer (IP); (ii) Outward Pointer (OP); (iii) Standalone Patch (SP); and (iv) Collaborative Patch (CP), as detailed in Sections 3 and 4. The findings suggest that integrating pointer and attention losses with the model loss is more beneficial than employing them individually. However, it also shows that only the pointer and attention losses can achieve significant adversarial success, provided  an adequate number and size of the patches. Previous attacks targeting dense dot-product attention, like Patch-Fool \cite{fu2022patch} and Attention-Fool \cite{lovisotto2022give} are concurrent works, and have been collectively labeled as Att. This method primarily draws from the latter work but also reflects elements from both strategies. Lastly, we also present the performance on clean samples to establish a baseline for comparison. Fig.~\ref{fig:bbox_CP} presents a comparison between the bounding boxes found in the ground truth, and those obtained from the clean and adversarial samples using Deformable DeTr. \\

\begin{figure}[t]
    \centering
    \includegraphics[width=\columnwidth ]{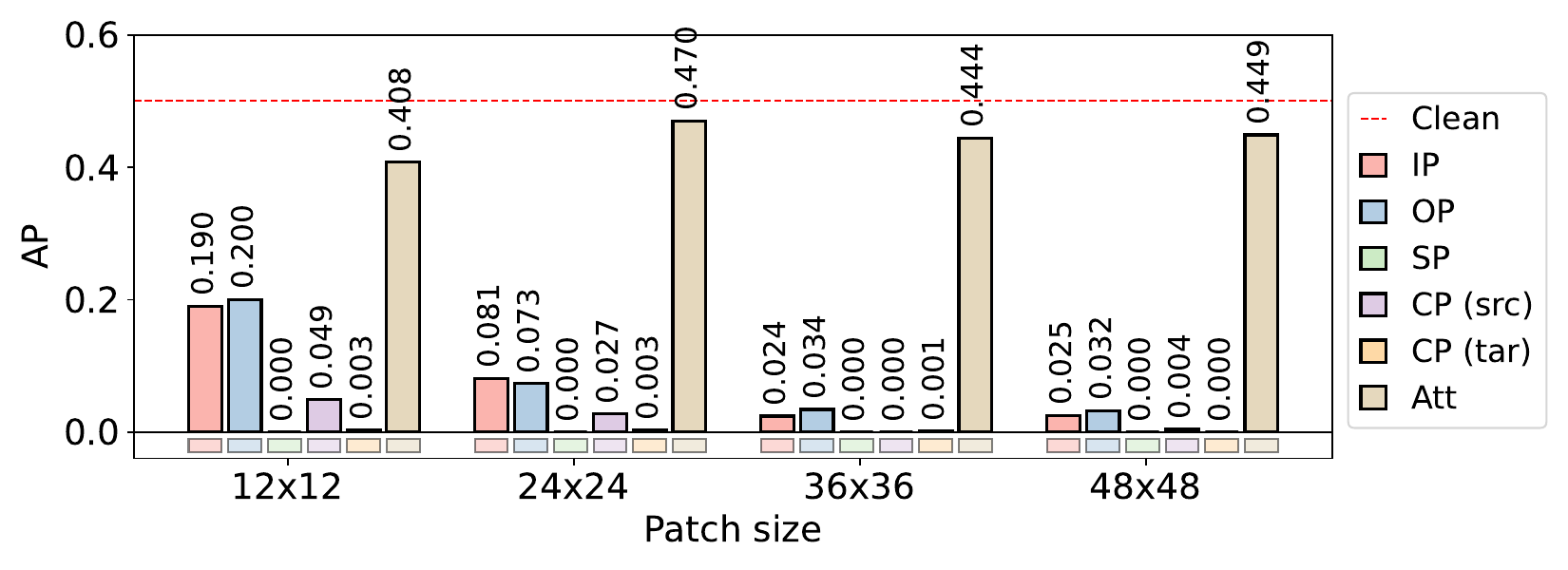}
    \caption{Impact of patch size on performance}
    \label{fig:patch_size}
\end{figure}

\noindent\textbf{Impact of Patch Size.} In this experiment, we carried out all four attacks to evaluate how the size of the patches influences their effectiveness. In IP, OP and SP attacks, we place a single patch on each views. In contrast, in CP attack we use two patches: a source and a target (the patch sizes are adjusted accordingly, ensuring that the total number of adversarially affected pixels was roughly equivalent to other attacks). The patch size is increased from 12x12 to 48x48, which is less than 0.4\% of the image size 768x768. Fig.~\ref{fig:patch_size} illustrates the Average Precision (AP) achieved with all four attacks. IP and OP attacks, which rely on pointer and attention losses, require larger patch sizes compared to SP and CP attacks that use combined losses.  Leveraging the adversarial model loss, CP and SP attacks more efficiently attack the system to reduces the AP to 0\% even with smaller patches. Lastly, the Att attack does not effectively reduce the performance even with larger patches. \\

\begin{figure}[t]
    \centering
    \includegraphics[width =\columnwidth ]{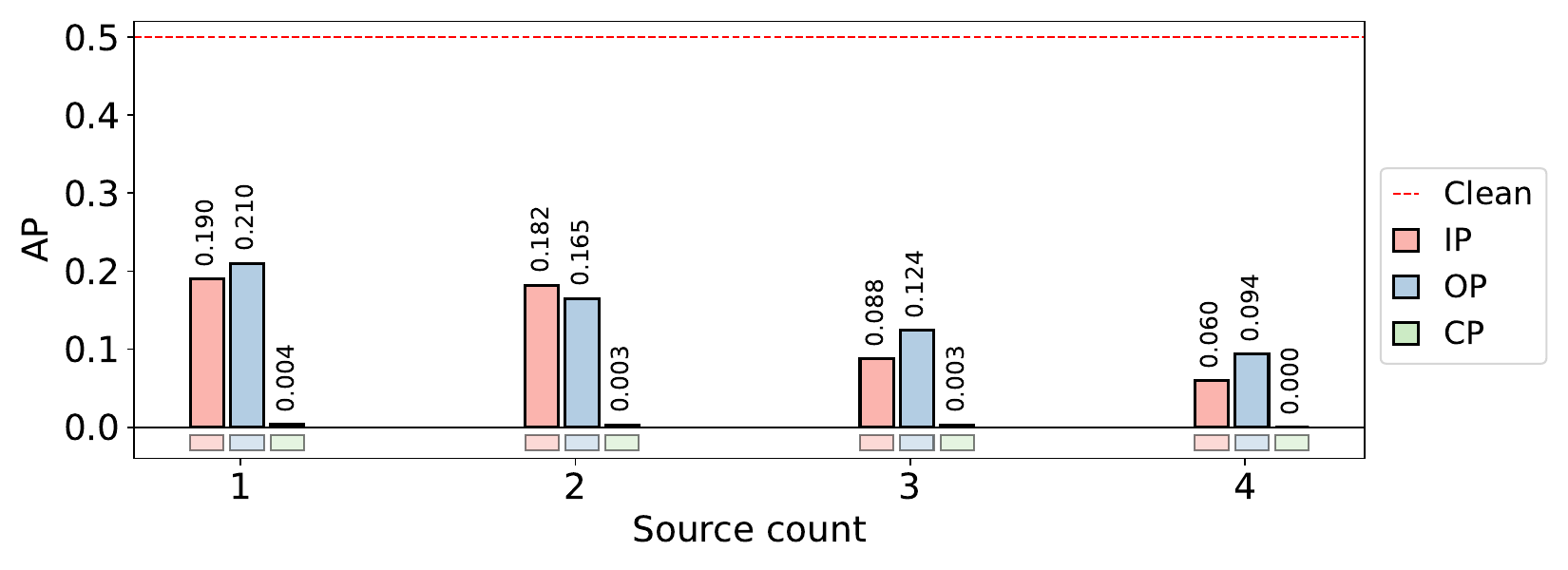}
    \caption{Impact of source patch count on performance}
    \label{fig:source_count}
\end{figure}

\noindent \textbf{Impact of Number of Source Patches}
In this experiment, we executed the attacks that have separate source patches, such as IP, OP, and CP attacks, to determine how the number of source patches affects attack performance. We omit the SP and Att attacks in this experiment, because in these attacks, either there is no source patch or the same patch acts as both source and target. For all the attacks used in this experiment, we employ a single target patch of size 12x12, while increasing the number of source patches (each of size 12x12) from 1 to 4. As depicted in Fig.~\ref{fig:source_count}, additional source patches improve the effectiveness of the attacks. \\

\begin{figure}[t]
    \centering
    \includegraphics[width =\columnwidth ]{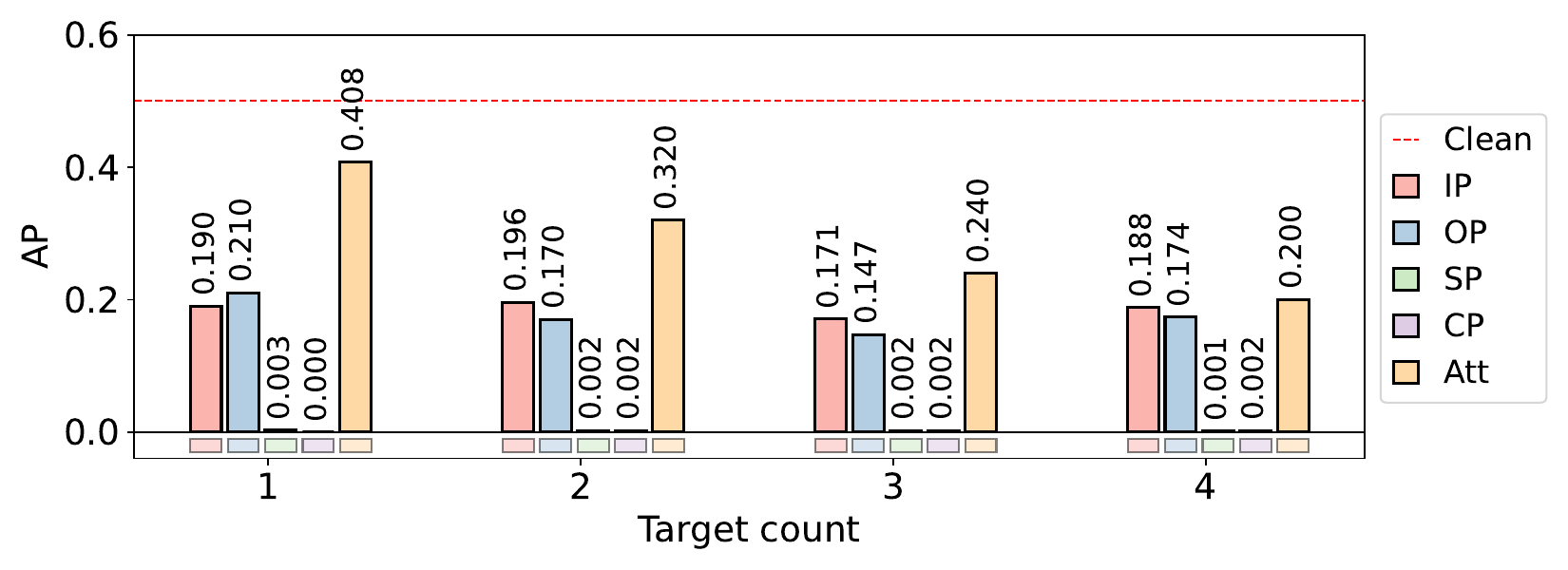}
    \caption{Impact of target patch count on performance}
    \label{fig:target_count}
\end{figure}

\noindent\textbf{Impact of Number Target Patches}
In this experiment, we carried out all four attacks to assess how the number of target patches influences the overall effectiveness of the attacks. We adjust the number of target points for IP and OP attacks, as target patches are not employed in these scenarios. For the other attacks, we used a single source patch of size 12x12, while increasing the number of target patches (each of size 12x12) from 1 to 4. As illustrated in Fig.~\ref{fig:target_count}, adding more target patches improves the attack success. This improvement can be attributed to the fact that additional target patches reduce the average distance from each pixel to its nearest target. Lastly, the Att attack benefits from additional target patches as they collectively have a greater impact on the pixels in their vicinity. \\



\begin{figure}[t]
    \centering
    \includegraphics[width =\columnwidth ]{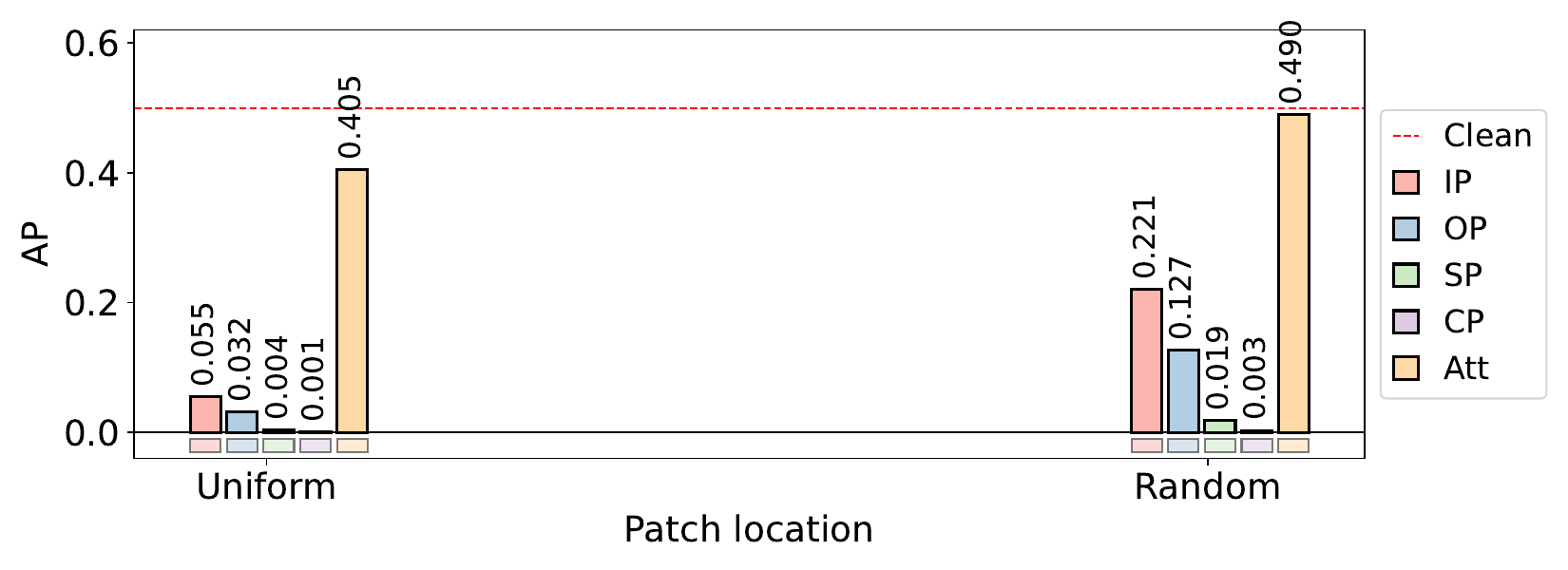}
    \caption{Impact of patch location on performance}
    \label{fig:patch_location}
\end{figure}

\noindent\textbf{Impact of Locations of Patches}
In this experiment, we carried out all four of our attacks to assess the impact of patch placement on the effectiveness of the attack. For all of the attacks, we either distributed the patches uniformly or randomly over the image or video frame. The source and target patches were distributed separately, and in the case of uniform distribution, we ensured non-overlap between them by applying a slight offset. As illustrated in Fig.~\ref{fig:patch_location}, uniform distribution performs better than random distribution, as it guarantees equal accessibility of the patches across the entire frame. \\



\begin{figure}[t]
    \centering
    \includegraphics[width =\columnwidth ]{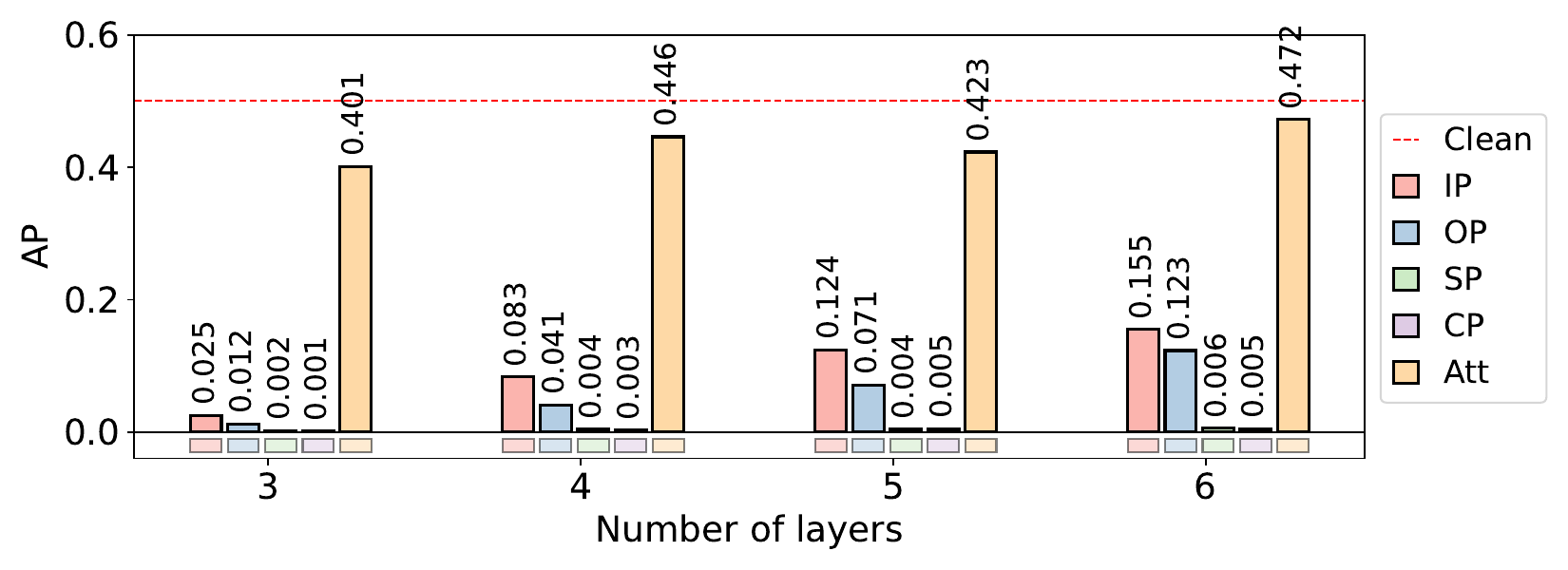}
    \caption{Impact of number of layers on performance}
    \label{fig:number_layers}
\end{figure}

\noindent\textbf{Impact of Number of layers}
In this experiment, we implemented all four of our attacks to evaluate the effect of the number of attention layers in the multi-headed DViT on attack efficacy. Intuitively, as the depth of the model increases, it learns more intricate relationships and dependencies among the tokens and, therefore, should be more difficult to attack. This observation is also reflected in our experiments (see Fig.~\ref{fig:number_layers}): as we increase the number of layers in our DViT architecture, the model becomes harder to attack, and vice versa. \\

\noindent \textbf{Impact of a Source Aggregator}
We explored a scenario where multiple source patches are placed at various locations. Instead of developing patches specific to each location, our aim was to design a universal patch effective from any of these locations. This approach has real-world implications for the practicality of these attacks. For instance, a single generic patch, like one printed on a T-shirt, could deceive the model if positioned in different areas of a scene or if the same patch appears at different locations in different cameras in a multi-view scenario.

\begin{figure}[t]
    \centering
    \includegraphics[width =\columnwidth ]{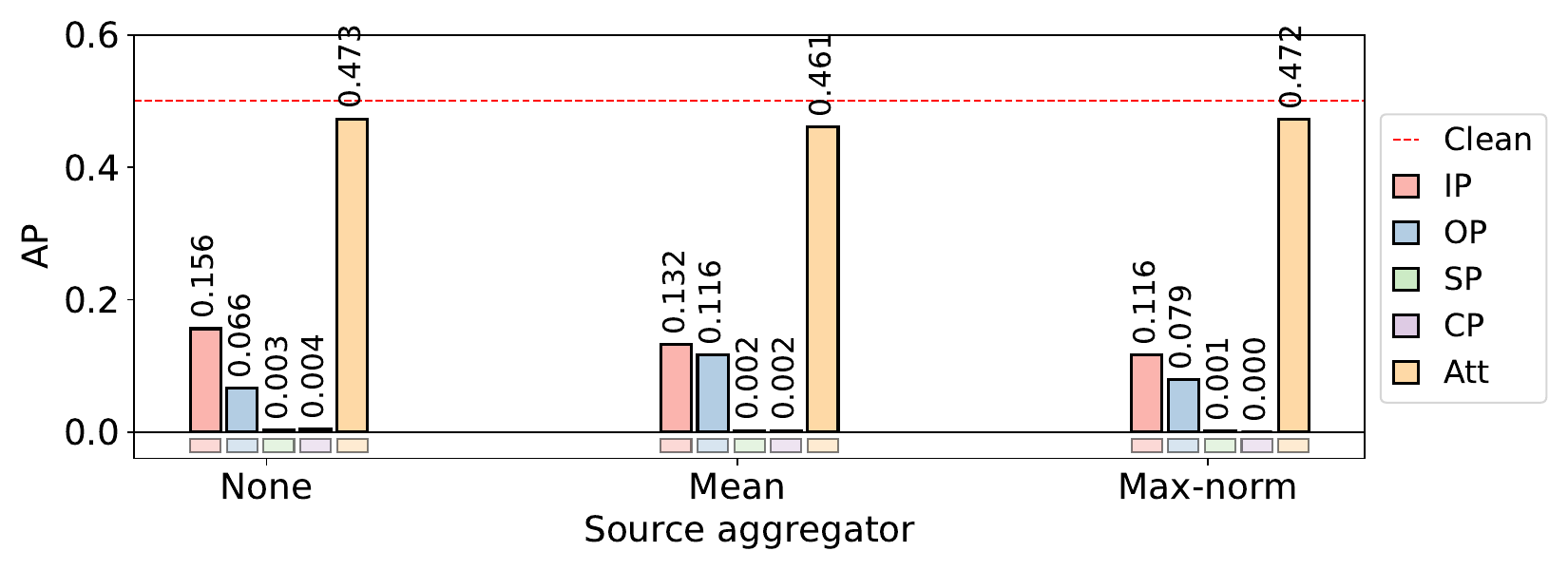}
    \caption{Impact of source aggregator on performance}
    \label{fig:source_aggregator}
\end{figure}

We explored two different approaches to create the patch: (1) using the mean of the gradients of the backpropagated loss over the patches; and (2) taking only the gradient that has the largest L2-norm.  Fig. \ref{fig:source_aggregator} shows the performance of the universal patch using the two aggregation approaches (Mean and Max-norm), for a multi-patch scenario.  In our experiments, multiple universal patches perform on par with having two distinct patches (shown as None on the figure), supporting that these attacks could transfer to real-world scenarios by leveraging universal patches. \\  

\subsection{Attack on Shadow Transformer}
\begin{figure}[t]
    \centering
    \includegraphics[width =0.78\columnwidth ]{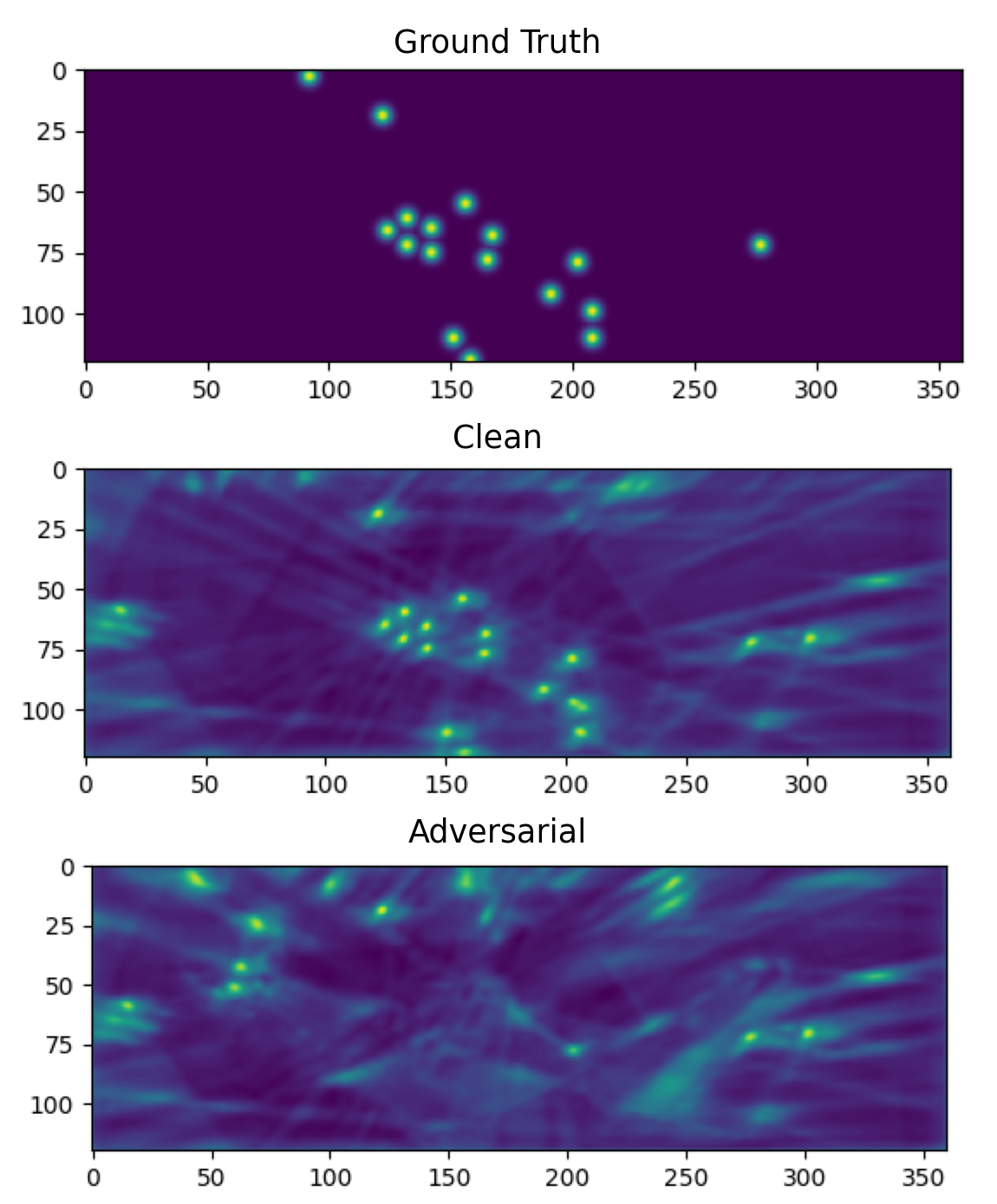}
    \caption{Heatmap comparison of the ground plane using the CP attack on MVDeTr: ground-truth vs. clean samples vs. adversarial samples. Notably, the object distribution is drastically different in the case of adversarial samples.}
    \label{fig:cp_hmap}
\end{figure}

\begin{figure*}[ht] 
    \centering
    \begin{minipage}{0.58\textwidth}
        \includegraphics[width=\linewidth]{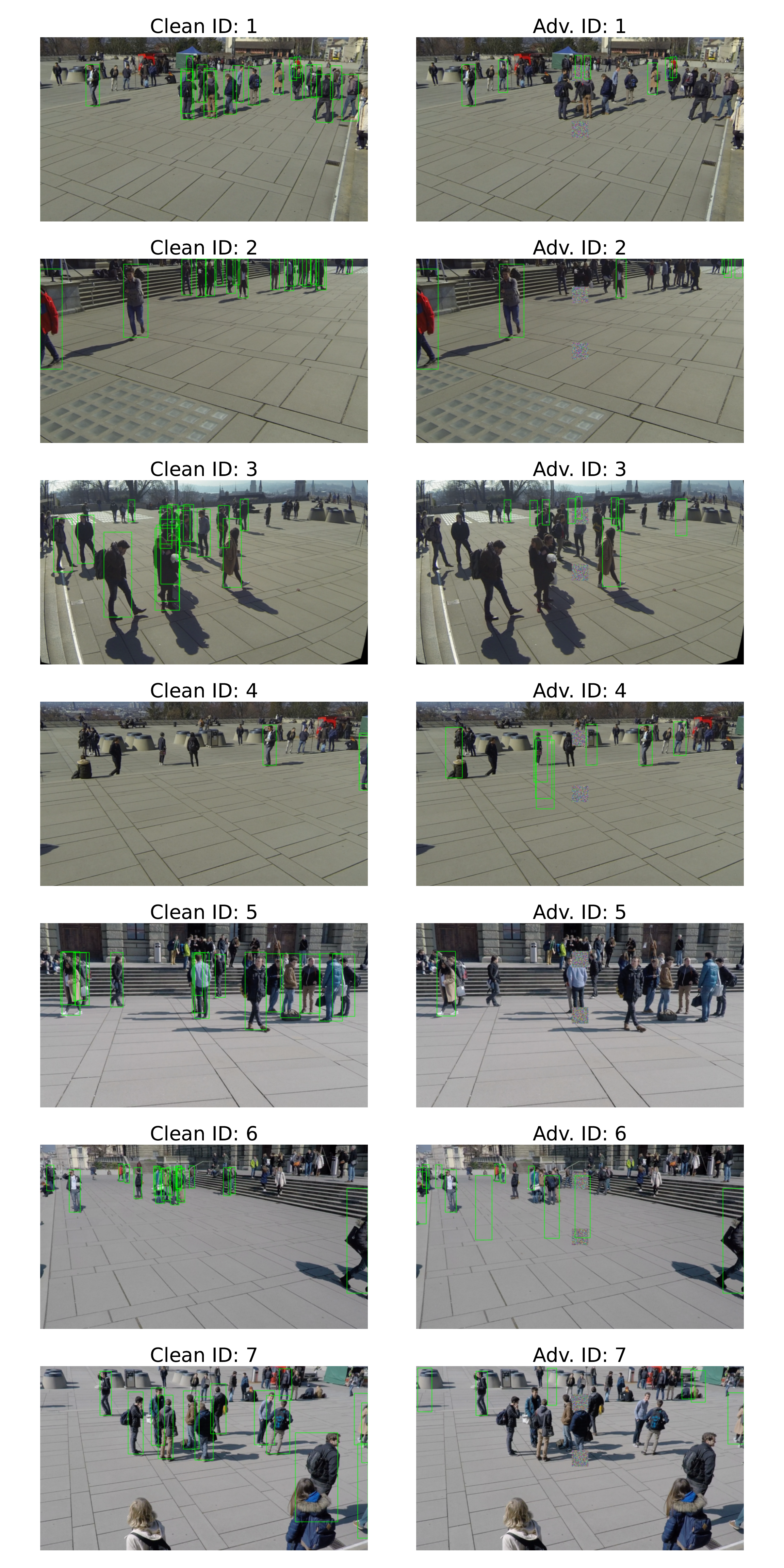}
        \caption{Bounding box comparison in each of the seven views using the CP attack on MVDeTr: clean samples vs. adversarial samples. Notably, the object detection performance is severely compromised (absent or false positives) in the case of adversarial samples.}
        \label{fig:bbox_mul_CP}
    \end{minipage}
\end{figure*}

We also extended our range of attacks to the Shadow Transformer \cite{hou2021multiview}, which is a key component of the Multi-view Object Detector (MVDeTr). In the MVDeTr framework, features from all views (seven in the Wildtrack dataset) are projected onto a shared ground plane. These features are then processed by a deformable transformer, named Shadow Transformer, which is specifically designed for multi-view systems. Unlike DeTr, where attention heads consists of pointers across different scales, in this implementation, they point across different views. To tailor these attacks for MVDeTr, we projected the locations of the source and target patches, and other attack related spatial information onto the ground place in order to incorporate them into the loss function. For MVDeTr, our focus was on SP and CP attacks. Fig.~\ref{fig:cp_hmap} presents a comparison between the heatmap of the ground plane found in the ground-truth and those obtained from the clean and adversarial samples using MVDeTr. Additionally, Fig~\ref{fig:bbox_mul_CP} presents a comparision between the bounding boxes in each of the seven views obtained from the clean and the adversarial samples.



\begin{figure}[t]
    \centering
    \includegraphics[width =\columnwidth ]{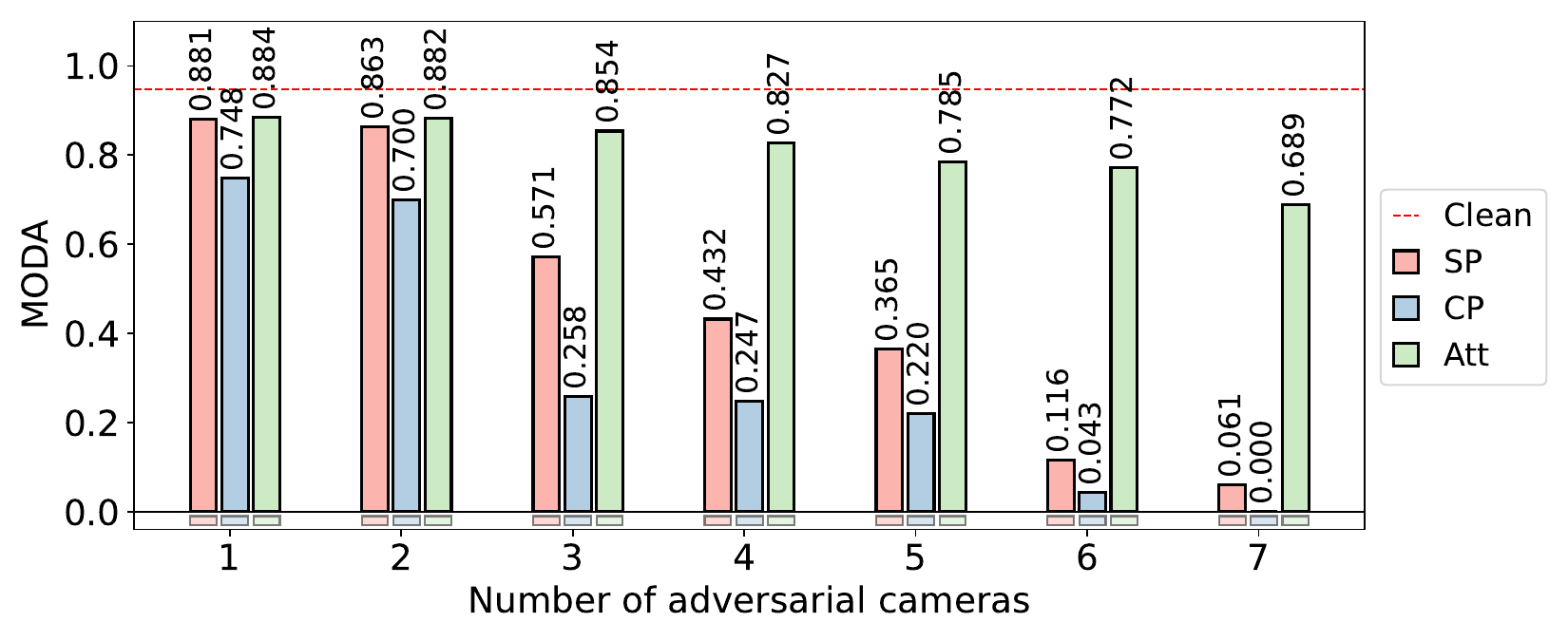}
    \caption{Impact of adversarial camera count on performance}
    \label{fig:num_cams_multiview}
\end{figure}

\noindent\textbf{Impact of Number of Adversarial Cameras.} In this experiment, we carried out both SP and CP attacks to evaluate how the number of adversarial cameras influences attack effectiveness. In both attacks, adversarial access was limited to a subset of the cameras. For the SP attack, each camera utilized a single patch of size 46x46. In contrast, the CP attack involved each camera having two patches, a source and a target (each of size 32x32), ensuring that the total number of adversarially affected pixels was roughly equivalent to that in the SP attack. As shown in Fig.~\ref{fig:num_cams_multiview}, the attacks were successful even when the adversarial access was limited to a subset of the cameras, since tokens (i.e., pixels) in one view have pointers to all views. Remarkably, with only 0.1\% of the total view area (across 7 views, each of size 1280x720), we were able to reduce the Multi-view Object Detection Accuracy (MODA) to 0\%. \\

\begin{figure}[t]
    \centering
    \includegraphics[width =\columnwidth ]{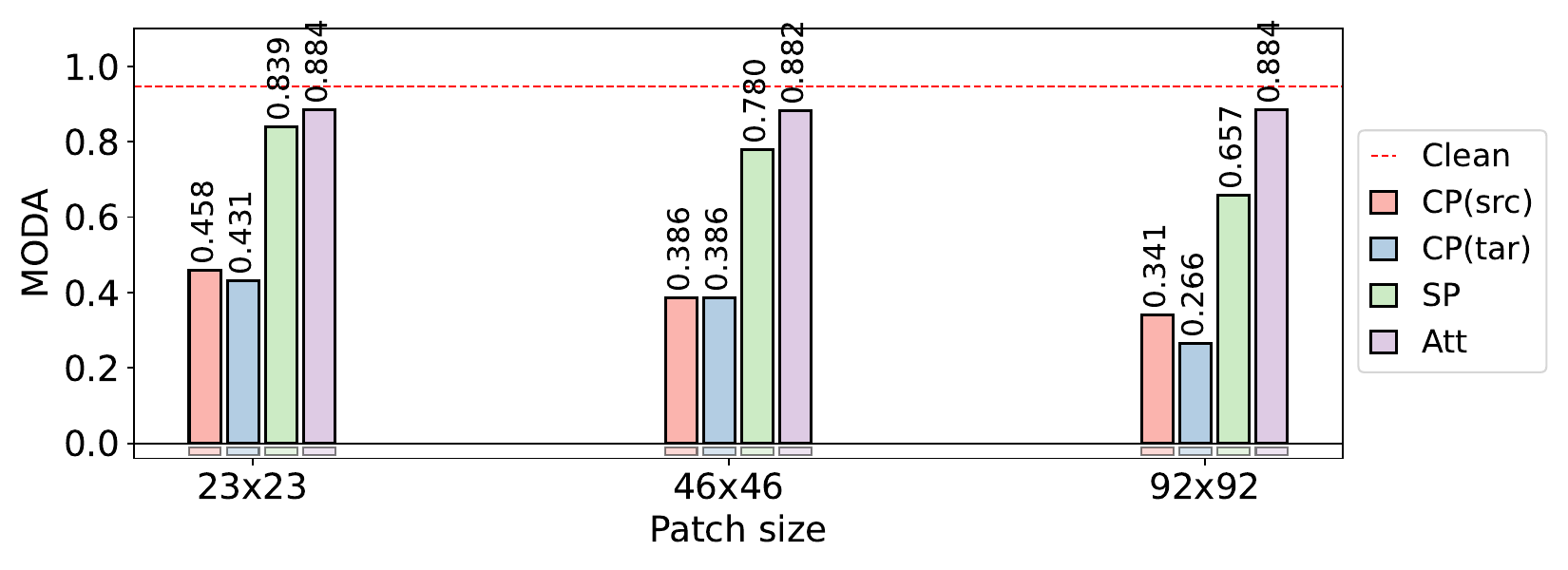}
    \caption{Impact of patch size on performance}
    \label{fig:num_size_multiview}
\end{figure}

\noindent\textbf{Impact of Patch Size.} In this experiment, we executed both SP and CP attacks to evaluate the effect of patch size on the success of the attacks. In the SP attack, each camera was equipped with a single patch, whereas, the CP attack involved two patches per camera, one source and one target. Furthermore, we explored two variations within the CP attack. In the CP(tar) setup, we increased the size of the target patch while keeping the source patch size constant, and in the CP(src) setup, we enlarged the source patch while the target patch size remained the same. As depicted in Fig.~\ref{fig:num_size_multiview}, the effectiveness of the attacks increased with the size of the patches. \\


\begin{figure}[t]
    \centering
    \includegraphics[width =\columnwidth ]{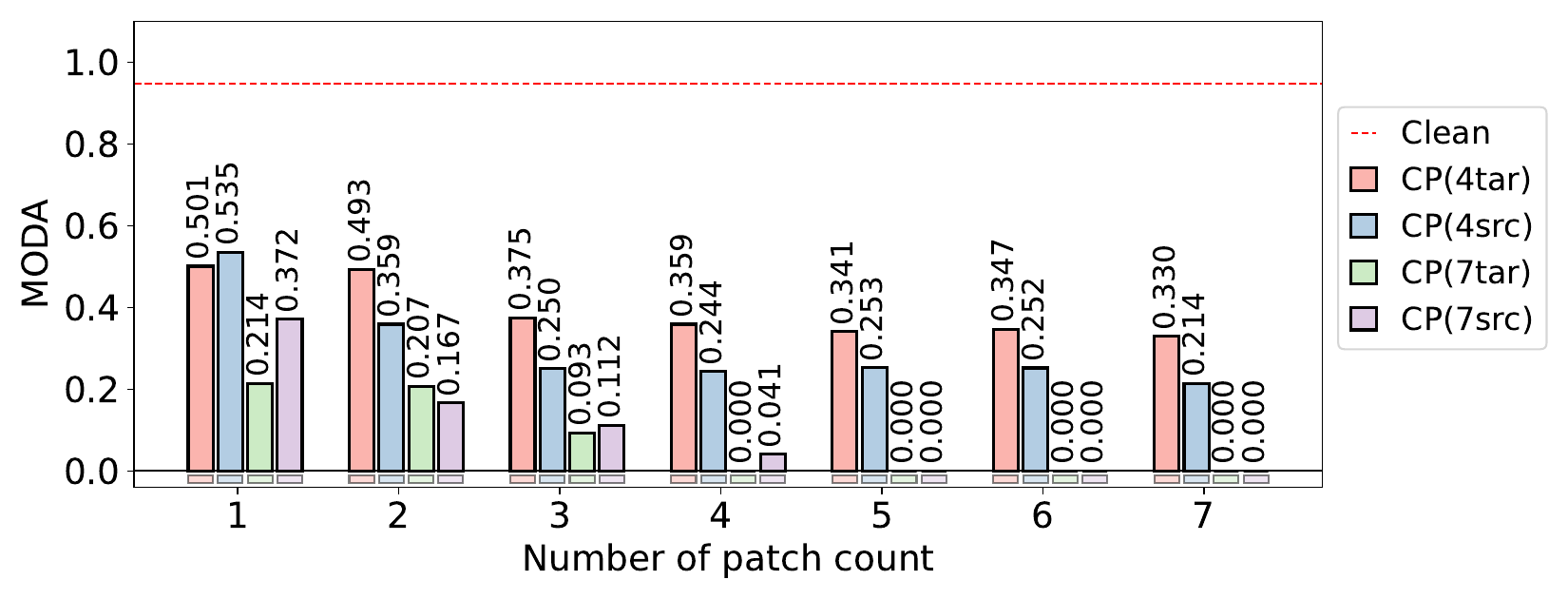}
    \caption{Impact of patch count on performance}
    \label{fig:num_patch_count_multiview}
\end{figure}

\noindent\textbf{Impact of Patch Count.} In this experiment, we implemented our CP attack to evaluate how the number of patches influences the effectiveness of the attack. In the CP(4tar) setup, we placed target patches in four views (from view 4 to 7). Subsequently, we systematically added one source patch to each of the views, starting with view 0. Similarly, in the CP(4src) setup, we first placed source patches in four views and then incrementally added target patches. For the CP(7tar) and CP(7src) setups, as opposed to four views, we placed target/source patches in all seven views respectively, before introducing patches of the other type. As shown in Fig.~\ref{fig:num_patch_count_multiview}, the attacks successfully compromised the system even when the source and target views were disjoint (for instance, CP(4tar) and CP(4src) with patches of other type in views 1 to 3). In CP(7tar) and CP(7src) setup, the attacks were extremely effective, reducing the Multi-view Object Detection Accuracy (MODA) to 0\%. 

\section{Related Work}
\label{sec:related}

Following the advent of attention-based transformers \cite{vaswani2017attention}, their use has become extensively popular in a diverse range of application areas. Initially designed for textual data, transformers have demonstrated exceptional promise in vision applications, and are being extensively explored. For instance, Vision Transformers (ViTs) have outperformed CNN-based models in object classification \cite{dosovitskiy2020image}, and DETR (Detection Transformer) has shown prowess in object detection \cite{carion2020end}. Subsequent efforts have focused on reducing the time complexity of the multi-headed dot-product attention used in conventional transformers \cite{zhu2020deformable}, enhancing their scalability and computational efficiency. Inspired by deformable convolution \cite{dai2017deformable} and dynamic convolution \cite{wu2019pay}, Zhu et al. introduced deformable mechanisms to the attention module of transformers, resulting in deformable detector-transformers (Deformable DeTrs) \cite{zhu2020deformable}. This innovation facilitated faster convergence and reduced runtime requirements, making Deformable DeTrs suitable for large feature spaces and multi-scale features. These advantages have driven the widespread adoption of deformable transformers in various domains, including the shadow transformer in the multi-view object detector MVDeTr \cite{hou2021multiview}, which models token relations across multiple views.

Adversarial attacks have proven to be a serious source of vulnerability for machine learning-based vision models \cite{goodfellow2014explaining}. Object classifiers and detectors have been shown to be vulnerable to adversarial attacks, and defenses as well as new attacks to bypass them continue to appear \cite{akhtar2018threat}. The robustness of transformer-based vision models, however, is still under investigation, with some initial studies suggesting they might be more resilient than CNNs. For example, Mahmood et al. found that attacks designed for CNNs are not easily transferable to transformers \cite{mahmood2021robustness}. Bhojanapalli et al. demonstrated that with sufficient data, transformers can be as robust as their ResNet counterparts \cite{bhojanapalli2021understanding}. Aldahdooh et al. suggest that this perceived robustness against adversarial attacks in transformers could be due to their tendency to capture global features \cite{aldahdooh2021reveal}. Other studies have shown that transformers focus more on low-frequency details, making them less susceptible to high-frequency patches \cite{benz2021adversarial, shao2021adversarial}.

Recent research, however, indicates that transformers might not be as robust as previously thought \cite{bai2021transformers, mahmood2021robustness}. For instance, Mahmood et al. demonstrated that an ensemble of transformers and CNNs in a white-box attack setting can be vulnerable \cite{mahmood2021robustness}. Wei et al. found that avoiding the dot-product attention during backpropagation can increase the transferability of attacks \cite{wei2022towards}. Recently, targeted attacks on the attention module of transformers have been successful in generating adversarial patches \cite{lovisotto2022give, fu2022patch}. Nevertheless, these approaches are specific to conventional transformer models with dot-product attention and are ineffective against learnable, data-driven sparse attention modules like those in deformable attention \cite{lovisotto2022give}.

\section{Concluding Remarks}

Deformable vision transformers (DViTs) exhibit intriguing properties from computational perspectives. These architectures not only achieve performance comparable to vision transformers (ViTs) but also are substantially more computationally efficient. The key to this efficiency lies in using a sparsified attention structure that adeptly captures the most important contextual relationships in a computationally streamlined manner, scalable to larger contexts (e.g., larger images, or systems reasoning about multiple views).

In this paper, we demonstrate that recent adversarial attacks developed against ViTs do not transfer effectively to DViTs. This inadequacy arises from the distinct nature of the attention mechanism in DViTs. Specifically, attention in DViTs is modeled as pointers to the most relevant keys for each query, along with their relative attention scores. Through an in-depth analysis of DViTs' security vulnerabilities, we introduce new attacks targeting these attention pointers.

Based on this principle, we propose a set of pointer-based attacks, each differing in how they manipulate the attention pointers. Our results indicate that these attacks significantly impair the performance of DViTs in both single-view and multi-view systems. The most effective strategy involves using an source patch to direct the pointers towards a target patch, aiming to compromise the model's global loss. This collaborative attack proves highly successful under various conditions, revealing the susceptibility of DViTs to adversarial attacks. Our findings broaden the threat model for vision transformers and we hope this will inform the development of more robust defenses against such attacks.

As vision systems grow in complexity, we anticipate that sparse attention transformers like DViTs will become increasingly pivotal. For instance, deformable vision transformers facilitate multi-view applications where the context includes several overlapping video streams, presenting a challenge to traditional transformer scalability. Vision language models and other systems also raise similar scalability issues. We aim to evaluate our attacks in the context of such systems in future research. Additionally, we plan to explore methods to reinforce the security of deformable transformers, including adversarial training and introducing constraints on the attention mechanism to prevent malicious manipulation.

{
    \small
    \bibliographystyle{ieeenat_fullname}
    \bibliography{main}
}



\end{document}